\documentclass[10pt,twocolumn,letterpaper]{article}

\usepackage[pagenumbers]{cvpr} 

\usepackage{graphicx}
\usepackage{amsmath}
\usepackage{amssymb}
\usepackage{booktabs}
\usepackage{algorithm}
\usepackage{algorithmic}
\usepackage{multicol}
\usepackage{multirow}
\usepackage{pifont}

\usepackage[pagebackref,breaklinks,colorlinks]{hyperref}

\usepackage[capitalize]{cleveref}
\crefname{section}{Sec.}{Secs.}
\Crefname{section}{Section}{Sections}
\Crefname{table}{Table}{Tables}
\crefname{table}{Tab.}{Tabs.}

\begin{document}

\title{Deep Incomplete Multi-view Clustering with Distribution Dual-Consistency Recovery Guidance}

\author{
{ Jiaqi Jin$^1$,
  Siwei Wang$^2$,
  Zhibin Dong$^1$,
  Xihong Yang$^1$,
  Xinwang Liu$^{1,}\thanks{Corresponding author}$ ~,
  En Zhu$^{1,*}$},
  Kunlun He$^3$\\
  $^1$School of Computer, National University of Defense Technology, Changsha, China\\
  $^2$Intelligent Game and Decision Lab, Academy of Military Sciences, Beijing, China\\
  $^3$Medical Big Data Research Center, Chinese PLA General Hospital, Beijing, China\\
  \texttt{jinjiaqi@nudt.edu.cn, wangsiwei13@nudt.edu.cn}
} 
\maketitle

\begin{abstract}
Multi-view clustering leverages complementary representations from diverse sources to enhance performance. However, real-world data often suffer incomplete cases due to factors like privacy concerns and device malfunctions. A key challenge is effectively utilizing available instances to recover missing views. Existing methods frequently overlook the heterogeneity among views during recovery, leading to significant distribution discrepancies between recovered and true data. Additionally, many approaches focus on cross-view correlations, neglecting insights from intra-view reliable structure and cross-view clustering structure. To address these issues, we propose BURG, a novel method for incomplete multi-view clustering with distri\textbf{B}ution d\textbf{U}al-consistency \textbf{R}ecovery \textbf{G}uidance. We treat each sample as a distinct category and perform cross-view distribution transfer to predict the distribution space of missing views. To compensate for the lack of reliable category information, we design a dual-consistency guided recovery strategy that includes intra-view alignment guided by neighbor-aware consistency and cross-view alignment guided by prototypical consistency. Extensive experiments on benchmarks demonstrate the superiority of BURG in the incomplete multi-view scenario.
\end{abstract}

\section{Introduction}\label{sec:intro}
In today's data-driven research landscape, multi-view clustering\cite{fang2023comprehensive, chen2022representation, xu2022self, lin2021multi, wang2024scalable, cui2024novel} aims to integrate information from different perspectives in an unsupervised manner, enhancing the predictive capability of the model\cite{li2018survey, liu2022simplemkkm}. However, in real-world scenarios, multi-view data is often incomplete due to high acquisition costs or privacy concerns\cite{liu2018late, xu2015multi, yu2024towards}. The missing information may hinder the discovery of underlying patterns and structures within the data, leading to inaccurate or unstable clustering. Therefore, the key to addressing this issue lies in how to utilize the available views to reasonably impute the missing ones\cite{wen2019unified, liu2024latent, pu2024adaptive}, thereby restoring the integrity of the data and maintaining the natural alignment of multi-view data\cite{liu2018late, li2021incomplete, liang2022incomplete}.

\begin{figure}[t]
\begin{center}{
    \centering
    \includegraphics[width=0.48\textwidth]{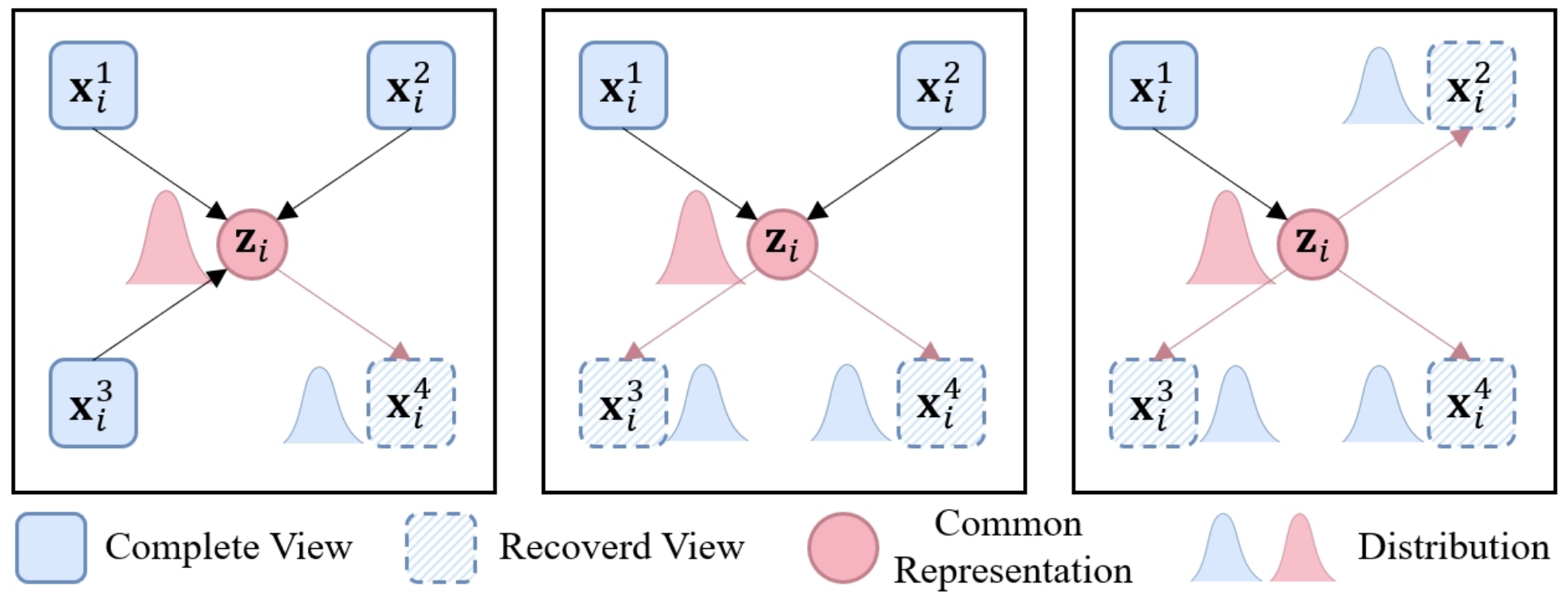}
    \caption{Basic idea of cross-view distribution transfer recovery. The consistent distribution obtained from the complete views is transferred to the missing views using view-specific flow models. The recovered representation is then input into the view-specific decoder to generate the recovered data.}
    \label{intro_fig}}
\end{center}
\vspace{-20pt}
\end{figure}

Currently, most mainstream deep incomplete multi-view clustering (DIMVC) methods first impute the missing instances in each view, and then perform clustering based on the complete and recovered views. These methods are commonly referred to as recovery-based DIMVC and can be broadly categorized into three types based on the recovery paradigm: GAN-based\cite{wang2021generative}, graph-structured\cite{wang2022incomplete, chao2024incomplete}, and information entropy-based methods\cite{lin2021completer, lin2022dual}.

However, existing recovery-based methods often assume that all views are similar in the feature space or can be directly mapped to the same feature space. This perspective overlooks the fact that different views may exhibit diverse feature distributions and data characteristics due to their varied sources, which is commonly referred to as the heterogeneity between views. Furthermore, these methods treat the recovery of missing view instances as independent across samples. While this paradigm may be suitable for tasks that seek one-to-one correspondences, the learned representations are not optimal for clustering, as the goal is to map $N$ representations into $K$ clusters. In other words, during the training process, these methods neglect the valuable guidance that various consistency structures—both within and across views—can provide for clustering.

Unlike previous methods for recovering missing data, we propose a deep incomplete multi-view clustering method with distri\textbf{B}ution d\textbf{U}al-consistency \textbf{R}ecovery \textbf{G}uidance, abbreviated as BURG, to mitigate the large discrepancy between the data recovered from complete views and the true data. The approximate recovery process for missing instances is outlined in Fig. \ref{intro_fig}. 

First, the complete views of each sample are processed through view-specific encoders and flow models to obtain representations that follow a prior distribution, which are then fused to form a consistent latent distribution. Finally, this representation undergoes the reverse process of the flow model corresponding to the view with missing instances, facilitating cross-view distribution transfer to recover the representation of the missing view. Specifically, due to the lack of manual annotations in an unsupervised setting, we initially treat each sample as independent and map its complete view features into a latent space that follows a Gaussian distribution through the forward process of the flow model. By leveraging the flow model's ability to explicitly measure distributional discrepancies, we minimize the gap between views, ensuring that the generated data is realistic and reliable. Additionally, without guidance from clustering structure information during recovery, the learned features may lack distinguishability. To address this, we introduce a dual consistency strategy to guide the recovery process: neighbor-aware consistency and prototypical consistency. These ensure the alignment process's reliability by enforcing consistency in cross-view neighbors and clustering assignments, respectively, enhancing the structural and clustering information within the features. 

Further details of the proposed BURG are shown in Fig. \ref{fig:frame}. The overall model framework is broadly divided into three modules: multi-view feature extraction (MFE), distribution transfer learning (DTL), and dual-consistency guided recovery (DGR). We summarize the major contributions of our work as follows,

\begin{itemize}
    \item We acknowledge the limitation of existing recovery-based DIMVC methods that neglect inter-view heterogeneity. In our proposed BURG, we employ a flow-based cross-view distribution transfer, which enables precise distribution transfer from observable views to missing views, thus minimizing the gap between the recovered and true data.
    \item We design a dual consistency strategy to guide both the data recovery process and representation learning. This strategy enriches the features with structural and clustering information, enhancing their discriminability and benefiting subsequent clustering tasks.
    \item Extensive experiments demonstrate the superiority of the proposed view-specific flow model-based distribution transfer recovery method, as well as the effectiveness of the dual consistency strategy.
\end{itemize}

\section{Related Work}\label{sec:related}
\subsection{Deep Incomplete Multi-view Clustering}\label{related:sub1}
Deep neural networks\cite{lecun2015deep, he2016deep} enhance clustering performance by effectively modeling associations across views\cite{guo2017improved, caron2018deep, niu2022spice, li2022twin, ren2024deep}, and various methods for deep incomplete multi-view clustering utilize this capability to tackle incomplete data. DIMVC methods can be classified into two main categories based on how they handle missing data: recovery-free and recovery-based approaches.

(1) Recovery-free methods learn representations solely from observable multi-view data, avoiding the complexities of imputing missing data.\cite{wang2020icmsc, feng2024partial, xue2024robust}. APADC\cite{xu2023adaptive} introduces an adaptive feature projection to bypass filling in missing data. DVIMC\cite{xu2024deep} employs the Product-of-Experts approach with consistency constraints to mitigate inter-view information imbalance, achieving a VAE-based recovery-free clustering method.

(2) Recovery-based methods focus on recovering missing data or its representations to mitigate the impact of incomplete data on clustering. These methods can be divided into three categories: \textit{i}) GAN-based methods, which use adversarial strategies to ensure generated data in missing views approximates observed data, such as CPM-Nets\cite{zhang2020deep} and GP-MVC\cite{wang2021generative}. \textit{ii}) Graph-structured methods, like CRTC\cite{wang2022incomplete}, SPCC\cite{dong2024subgraph} and ICMVC\cite{chao2024incomplete}, transfer graph structures from complete views to fill in missing data. \textit{iii}) Information entropy-based methods, including COMPLETER\cite{lin2021completer} and its variant DCP\cite{lin2022dual}, minimize conditional entropy through dual prediction to recover missing data. DIVIDE\cite{lu2024decoupled} follows a similar principle, using cross-view decoders to recover samples across views.


\begin{figure*}[!ht]
    \centering
    \includegraphics[width=0.99\textwidth]{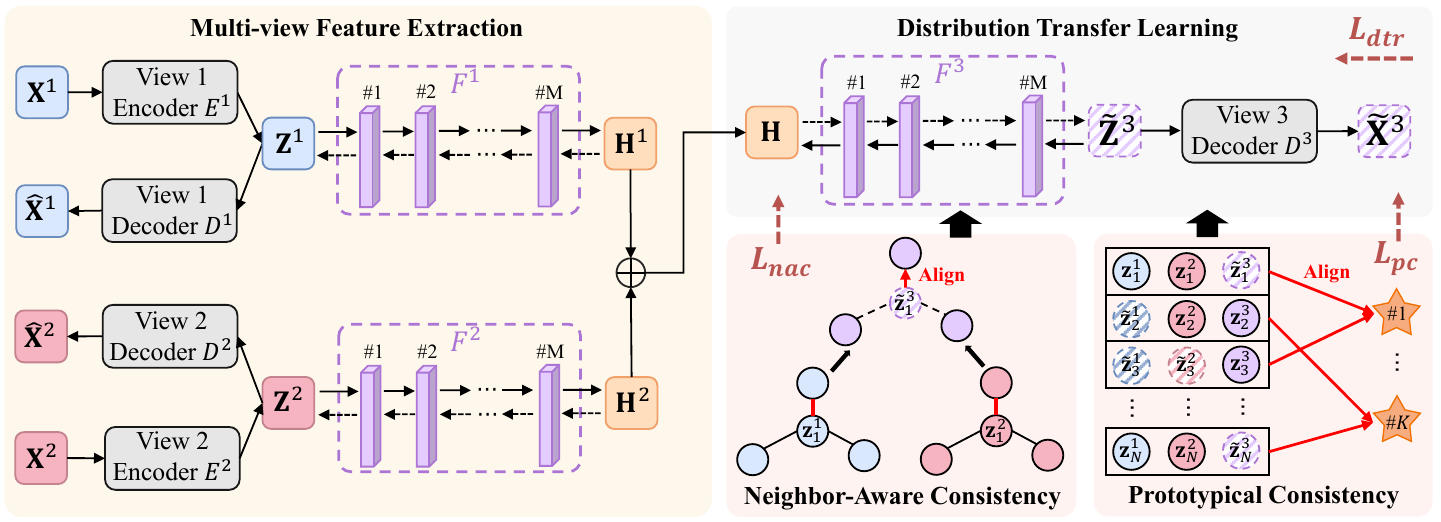}
    \caption{The framework of BURG. We use three views as an example, where the first and second views are complete, while the third is missing. As shown, BURG consists of a joint optimization of three modules: multi-view feature extraction(MFE), distribution transfer learning(DTL), and dual-consistency guided recovery(DGR). The two types of consistency are neighbor-aware and prototypical consistency. In DTL, the common embedding $\mathbf{H}$ is formed by combining $\mathbf{H}^1$ and $\mathbf{H}^2$, which are obtained through view-specific forward flows. Subsequently, $\mathbf{H}$ is passed through reverse flow $(F^3)^{-1}$ to generate the latent representation $\tilde{\mathbf{Z}}^3$ of the missing view. Furthermore, DGR provides essential local intra-view structure and global inter-view clustering structure throughout the training process to ensure the discriminability of the recovered representation.}
    \label{fig:frame}
\end{figure*}

\subsection{Normalizing Flow Model}\label{related:sub2}
Unlike generative models like GANs\cite{goodfellow2014generative}, Normalizing Flows\cite{dinh2014nice, kingma2018glow} explicitly model the probability distribution of data, allowing for precise likelihood estimation. Flow-based generative models map complex data distributions to simpler ones, such as Gaussian or logistic distributions, through a series of invertible and differentiable transformations. In practical applications, Normalizing Flow is widely used in areas such as image and text generation\cite{abdal2021styleflow}, anomaly detection\cite{hirschorn2023normalizing} and image restoration\cite{wang2022low}.

Mathematically, let $\mathbf{z} \sim p_Z(\mathbf{z})$ be a latent variable that obeys a simple distribution, and $\mathbf{x}$ is a complex data sample. A flow model defines a reversible transformation from $\mathbf{z}$ to $\mathbf{x}$ via a mapping function $F$. Using the variable transformation formula, the probability distribution $p(\mathbf{x})$ of $\mathbf{x}$ can be expressed as follows,
\begin{equation}
\label{flow_origin}
 p_X(\mathbf{x}) = p_Z(F^{-1}(\mathbf{x})) \left| \det\left( \frac{\partial F^{-1}(\mathbf{x})}{\partial \mathbf{x}} \right) \right|,
\end{equation}
where $F^{-1}$ represents the inverse transformation, and $\det\left( \frac{\partial F^{-1}(\mathbf{x})}{\partial \mathbf{x}} \right)$ denotes the determinant of the Jacobian matrix of the inverse transformation.

Flow-based models are highly promising for IMVC because they can model complex distributions while providing precise likelihood estimates. By using invertible transformations, flow network can learn dependencies between different views and recover missing entries.

\section{Method}\label{sec:method}
In this section, we will first sequentially introduce the three modules in BURG, as shown in Fig. \ref{fig:frame}, and then present the objective function and optimization process of the model.

\textbf{Notations}. We denote the multi-view data as $\{\mathbf{X}^{v} \}^V_{v=1}$, where $V$ is the number of views. Each view $\mathbf{X}^{v} = \{\mathbf{x}^{v}_1, \mathbf{x}^{v}_2, \cdots, \mathbf{x}^{v}_N\} \in \mathbb{R}^{N \times d_v}$ consists of $N$ instances with dimensionality $d_v$. The set $\mathbf{W}=\{\mathbf{w}^{v}\}^V_{v=1}$ indicate the presence of missing instances, where $\mathbf{w}^{v}=\{w^{v}_1, w^{v}_2, \cdots, w^{v}_N\} \in \mathbb{R}^N$. Concretely, 
\begin{equation}
\label{missmat}
w^{v}_i=
\begin{cases}
1, & \text{if $i$-th sample is complete in $v$-th view}\\
0, & \text{otherwise}
\end{cases}. 
\end{equation}

The encoder, decoder and flow network for $v$-th view are denoted by $E^v$, $D^v$ and $F^v$, respectively.

\subsection{Multi-view Feature Extraction}\label{sec31}
\subsubsection{Common and View-Specific Reconstruction}\label{sec311}
For the $v$-th view, the encoder-decoder pair $E^v$ and $D^v$ is utilized to project different views into a common subspace, with the resulting high-level representation denoted as $\mathbf{Z}^{v}$. The projection of encoder is expressed as $E^v(\mathbf{X}^{v}): \mathbf{X}^{v} \in \mathbb{R}^{N \times d_v} \to \mathbf{Z}^{v} \in \mathbb{R}^{N \times d} $, and feature reconstruction of decoder is formulated as $D^v(\mathbf{Z}^{v}): \mathbf{Z}^{v} \in \mathbb{R}^{N \times d} \to \hat{\mathbf{X}}^{v} \in \mathbb{R}^{N \times d_v} $, where $d$ represents the dimension of the common subspace.

To enhance recovery and alignment in the next modules, we input both the view-specific and common representations into $D^v$ to reconstruct complete instances. The reconstruction loss can be expressed as follows:
\begin{equation}
\label{Lrec}
\begin{split}
\ell^v_i =   \left\| \mathbf{x}_i^{v} - D^v(\mathbf{z}_i^{v})  \right\|_{2}^2  +  \left\| \mathbf{x}_i^{v} - D^v(\mathbf{z}_i)  \right\|_{2}^2,
\end{split}
\end{equation}
where the first and second terms correspond to decoding the view-specific/common representation $\mathbf{z}_i^{v} / \mathbf{z}_i$ of $i$-th sample, coefficient $w^{v}_i$ ensures that reconstruction is performed only on complete instances. The calculation for the two latent representations in Eq. (\ref{Lrec}) are as follows,
\begin{equation}
\label{z_obtain}
\begin{split}
\mathbf{z}_i^{v}=E^v(\mathbf{x}_i^{v}), \  & \mathbf{z}_i = \frac{1}{\sum\nolimits_{v=1}^V w^{v}_i} \sum\limits_{v=1}^V w^{v}_i \mathbf{z}^v_i.
\end{split}
\end{equation}
Based on Eqs. (\ref{Lrec}) and (\ref{z_obtain}), the information reduction during the reconstruction process can be minimized, resulting in effective subspace representations.

\subsubsection{Flow-based Distribution Learning}\label{sec312}
This section formalizes the forward generation process of the flow model, where the embedding $\mathbf{z}^v_i$ obtained from the encoder $E^v$ is input into $F^v$ to produce a representation $\mathbf{h}^v_i$ that follows a standard Gaussian distribution, expressed as $\mathbf{h}^v_i \leftarrow F^v(\mathbf{z}^v_i) \sim\mathcal{N}(\mathbf{0}, \mathbf{I})$.
Our flow-based generative network $F^v$ consists of $M$ coupling layers and a scaling layer, represented as $F^v=f^v_1 \circ f^v_2 \circ \cdots \circ f^v_M \circ s^v$. Unlike traditional Normalizing Flow, we incorporate affine transformations into each coupling layer to better adapt the model to complex data distributions.

First, we split $\mathbf{z}^v_i \in \mathbb{R}^d$ into two parts along its dimensions, i.e., $\mathbf{z}_i^v = [ \mathbf{a}^\top ; \mathbf{b}^\top ]^\top$. The output of the $m$-th affine coupling layer is denoted as $\mathbf{h}^v_{i,m} = [ \mathbf{a}_m^\top ; \mathbf{b}_m^\top ]^\top$. The nonlinear transformation applied by the coupling layer varies with the parity of $m$. Concretely, 
\begin{equation}
\label{coupling}
\begin{bmatrix} \mathbf{a}_m \\ \mathbf{b}_m \end{bmatrix} = 
\begin{cases}
\begin{bmatrix} \mathbf{a}_{m-1} \odot e^{s^m_c(\mathbf{b}_{m-1})} + t^m_c(\mathbf{b}_{m-1}) \\ \mathbf{b}_{m-1} \end{bmatrix}, \text{$m$ is odd} \\
\\
\begin{bmatrix} \mathbf{a}_{m-1} \\ \mathbf{a}_{m-1} \odot e^{s^m_c(\mathbf{b}_{m-1})} + t^m_c(\mathbf{b}_{m-1}) \end{bmatrix},\text{$m$ is even}
\end{cases}
\end{equation}
where $s^m_c(\cdot)$ and $t^m_c(\cdot)$ are the scaling and translation functions within the coupling layer, and $\odot$ denotes element-wise multiplication.

In flow model, Jacobian determinant of the distribution transformation can be computed from the output of the coupling layer. For the $m$-th coupling layer, the Jacobian matrix $J_m$ can be expressed in the following block form,
\begin{equation}
\label{Jacobian}
J_m = 
\begin{cases}
\begin{bmatrix}
diag(e^{s^m_c(\mathbf{b}_{m-1})}) & \frac{\partial t^m_c(\mathbf{b}_{m-1})}{\partial \mathbf{b}_{m-1}} \\
\mathbf{0} & \mathbf{I}
\end{bmatrix}, \text{$m$ is odd} \\
\\
\begin{bmatrix}
\mathbf{I} & \mathbf{0} \\
\frac{\partial t^m_c(\mathbf{a}_{m-1})}{\partial \mathbf{a}_{m-1}} & diag(e^{s^m_c(\mathbf{a}_{m-1})})
\end{bmatrix},\text{$m$ is even}
\end{cases}.
\end{equation}

The flow model incorporates a scaling layer after coupling layers to control data scale and enhance expressive capability. The transformation process of the scaling layer is as follows,
\begin{equation}
\label{scaling}
\mathbf{h}^v_i = \exp(\mathbf{s}^v_\theta) \odot [ \mathbf{a}_M^\top ; \mathbf{b}_M^\top ]^\top ,
\end{equation}
where $\mathbf{s}^v_\theta$ represents the learnable parameters of the scaling layer. Additionally, the Jacobian determinant of the scaling layer is simply the sum of the elements of $\mathbf{s}^v_\theta$.

According to the definition in Eq. (\ref{flow_origin}), taking the logarithm of both sides yields the following log-likelihood of the difference between the distributions before and after the transformation,
\begin{equation}
\label{log_p}
\log p(\mathbf{z}^v_i) = \log p(\mathbf{h}^v_i) + \log \left| \det J \right|,
\end{equation}
where $\left| \det J \right|$ is the sum of the products of the Jacobian determinants of all coupling layers, along with the influence of the scaling layer. In our model, we assume that the output $\mathbf{h}^v_i$ of the forward process of the flow model follows a standard Gaussian distribution. Substituting the corresponding probability density function into Eq. (\ref{log_p})yields the following expression,
\begin{equation}
\label{logp_detail}
\begin{aligned}
\log p(\mathbf{z}^v_i) = & -\frac{1}{2} (d\log2\pi+\left\| \mathbf{h}_i^{v} \right\|_{2}^2) \\
& + \log \left| \det J_s \right| + \sum\limits_{m=1}^{M} \log \left| \det J_m \right|.
\end{aligned}
\end{equation}

Eq. (\ref{logp_detail}) can explicitly and accurately describe the data distribution, thereby facilitating an effective initialization for our view-specific flow model.

\subsection{Distribution Transfer Learning}\label{sec32}
As shown in Fig. \ref{fig:frame}, the main goal of DTL module is to transfer the distribution of the common representation obtained from the observable views to the missing view based on the view-specific flow model. This enables the accurate estimation of missing data representations that are highly consistent with the real data.

Assuming that the $v$-th view instance of the $i$-th sample is missing while the others are observable, the rough process of recovering the missing representation $\tilde{\mathbf{z}}^v_i$ from the consistent embedding $\mathbf{h}_i$, which follows a standard Gaussian distribution, through $(F^v)^{-1}$ is as follows,
\begin{equation}
\label{flow_reverse}
\mathbf{h}_i \stackrel{(s^v)^{-1}}{\longrightarrow} \mathbf{h}^v_{i,M} \stackrel{(f^v_M)^{-1}}{\longrightarrow} \cdots \stackrel{(f^v_2)^{-1}}{\longrightarrow} \mathbf{h}^v_{i,1} \stackrel{(f^v_1)^{-1}}{\longrightarrow} \tilde{\mathbf{z}}^v_i,
\end{equation}
where $\mathbf{h}_i$ is derived from all available views of the $i$-th sample, denoted as $\{\mathbf{z}^{v’}_i | v’=1,\cdots,V, v’ \ne v \}$. The process of obtaining the consistent distribution representation $\mathbf{h}_i$ is as follows,
\begin{equation}
\label{get_hi}
\mathbf{h}_i = \frac{1}{\sqrt{V-1}} \sum\limits_{v'=1, v' \ne v}^{V} F^{v'} (\mathbf{z}^{v'}_i) \sim\mathcal{N}(\mathbf{0}, \mathbf{I}),
\end{equation}
based on the scaling properties of the mean vector and covariance matrix of the Gaussian distribution, it can be straightforwardly derived that $\mathbf{h}_i$ still follows a standard Gaussian distribution.

Therefore, the error in cross-view distribution transfer within the DTL module can be quantified by the reconstruction loss between the recovered representation and the true representation $\mathbf{z}^v_i$, which is formally expressed as follows,
\begin{equation}
\label{loss_dtl}
\begin{aligned}
\mathcal{L}_{dtl} = \sum_{i=1}^{N} \sum_{v=1}^{V} &  \left( 
\prod_{v'=1}^{V} w^{v'}_i \right) \left\| (F^v)^{-1}(\mathbf{h}_i) - \mathbf{z}_i^v  \right\|_{2}^2
 \\
&  + w^v_i (\ell^v_i - \log p(\mathbf{z}^v_i)),
\end{aligned}
\end{equation}
where the first loss term is used to minimize the error in cross-view distribution transfer learning, while the second term is aimed at preserving the multi-view feature extraction capability learned from the MFE module, thereby preventing catastrophic forgetting in the deep model.

\subsection{Dual-Consistency Guided Recovery}\label{sec33}

\subsubsection{Neighbor-Aware Consistency}\label{sec331}
Heterogeneity is reflected in the feature distributions and data characteristics between views, representing a global property. In contrast, similarity between views manifests in the local cross-view structure, representing a local neighbor property.
In this work, this local similarity is embodied in the consistency of cross-view nearest neighbor relationships. When the $v$-th view of a sample is missing, we transfer the nearest neighbor relationships from other views to the $v$-th view, where the recovered representation $\tilde{\mathbf{z}}^v_i$ resides. The process of identifying cross-view nearest neighbor relationships is as follows,
\begin{equation}
\label{nearest}
n_a = \mathop{\min}_{v' \in \{1, \cdots ,V\} \atop v' \ne v} \sigma (\Omega^{v'} (\tilde{\mathbf{z}}^v_i), \mathbf{z}^{v'}_i),
\end{equation}
where $\sigma(\cdot, \cdot)$ denotes the similarity measure function, and $\Omega^{v'} (\tilde{\mathbf{z}}^v_i)$ represents the nearest neighbor representation of $\tilde{\mathbf{z}}^v_i$ in $v'$-th view.

Subsequently, the nearest neighbor relationship is are transferred to the $v$-th view, and $\tilde{\mathbf{z}}^v_i $ is encouraged to approach its nearest neighbor representation within this view. The loss associated with this process can be formalized as follows,
\begin{equation}
\label{Lnac}
\mathcal{L}_{nac} = \sum\limits_{i=1}^{N} \sum\limits_{v=1}^V (1 - w^{v}_i) \sigma (\tilde{\mathbf{z}}^v_i, \mathbf{z}^v_{n_a}).
\end{equation}

\subsubsection{Prototypical Consistency}\label{sec332}
Based on the assumption that the clusters to which samples belong across different views should be consistent, we propose prototypical consistency. This assumption is essential in multi-view clustering. First, we perform $k$-means on the consistent representations obtained from the encoder for the complete views, resulting in a set of consistent prototypes $\{\mathbf{c}_k\}^{K}_{k=1}$ for all samples. Subsequently, the normalized $K$-dimensional cluster assignment is obtained from the following formula,
\begin{equation}
\label{softmax_p}
p(\mathbf{z}^v_i, \mathbf{c}_k) = \frac{\exp{(\sigma(\mathbf{z}^v_i, \mathbf{c}_k) / \tau)}} {\sum\nolimits_{{k'}=1}^{K} \exp{(\sigma(\mathbf{z}^v_i, \mathbf{c}_{k'}) / \tau)}} ,
\end{equation}
where temperature coefficient $\tau$ is fixed at 1.0 in this work. For simplicity, we abbreviate $p(\mathbf{z}^v_i, \mathbf{c}_k)$ as $p^v_{i,k}$ in the following formulas.

In unsupervised scenario, the recovered views cannot be more reliable than complete ones. Therefore, we derive the prototype assignment for the entire sample using only the complete views,
\begin{equation}
\label{pred_y}
\tilde{y}_i = {\underset {k \in \{1, \cdots ,K\}} { \operatorname {arg\,max} } \, \left( \frac{1}{\sum\nolimits_{v=1}^V w^{v}_i} \sum\limits_{v=1}^V p^v_{i,k} \right)}.
\end{equation}

Finally, we align both the complete and recovered views of each sample towards its consistent prototype, resulting in the following prototypical consistency alignment loss,
\begin{equation}
\label{Lpc}
\mathcal{L}_{pc} = - \sum\limits_{i=1}^{N} \left[ \sum\limits_{v=1}^V \sum\limits_{k=1}^K \left( \tilde{y}_{i,k} \log p^v_{i,k} + \gamma p^v_{i,k} \log p^v_{i,k}    \right) \right],
\end{equation}
where $\tilde{y}_{i,k} \in \tilde{\mathbf{y}}_i$ and $\tilde{\mathbf{y}}_i = {\rm{one\_hot}}(\tilde{y}_i)$.
In Eq. (\ref{Lpc}), the latter of the two loss terms is Shannon entropy, which serves as a regularization term to prevent category collapse, ensuring that not all samples are assigned to the same cluster. The parameter $\gamma$ is used to balance the strength of the regularization term.

\begin{algorithm}[t]
    \caption{The Optimization of BURG}
    \label{algorithm:BURG}
    \renewcommand{\algorithmicrequire}{\textbf{Input:}} 
    \renewcommand{\algorithmicensure}{\textbf{Output:}} 
    \begin{algorithmic}[1]
        \REQUIRE 
        Dataset $\{\mathbf{X}^{v} \}^V_{v=1}$ of size $N$; number of clusters $K$; networks $\{E^v, D^v, F^v\}^V_{v=1}$; max epoch of three stage $E_1, E_2, E_3$; trade-off parameters $\alpha$ and $\beta$.
        \ENSURE 
        The predicted clustering results $\hat{Y}$.
        
        \FOR{$e = 1$ to $E_1$}
            \STATE Update $\{E^v, D^v\}^V_{v=1}$ with Eq. (\ref{Lrec})
            \STATE Update $\{F^v\}^V_{v=1}$ with Eq. (\ref{logp_detail})
        \ENDFOR
    
        \FOR{$e = 1$ to $E_2$}
            \STATE Obtain $\mathbf{h}_i$ with Eq. (\ref{get_hi})
            \STATE Train the entire network with Eq. (\ref{loss_dtl})
        \ENDFOR
        
        \FOR{$e = 1$ to $E_3$}
            \STATE Obtain the $n_a$ and $\tilde{y}_i$ with Eq. (\ref{nearest}) and Eq. (\ref{pred_y})
            \STATE Train the entire network with Eq. (\ref{obj_func})
        \ENDFOR
        
    \end{algorithmic}
\end{algorithm}

\subsection{Objective Function and Optimization}\label{sec34}
Our overall optimization objective consists of three components: the distribution transfer learning loss, the neighbor-aware consistency loss, and the prototypical consistency loss. In general, the objective loss function of BURG is formulated as follows,
\begin{equation}
\label{obj_func}
\mathcal{L} = \mathcal{L}_{dtl} + \alpha \mathcal{L}_{nac} + \beta \mathcal{L}_{pc},
\end{equation}
where $\alpha$ and $\beta$ are balanced hyperparameters used to adjust the influence of two types of consistency.

The whole optimization procedure is listed in Algorithm \ref{algorithm:BURG}. After the entire optimization, we concatenate the missing view representations recovered by BURG with the complete views and then perform $k$-means to obtain the final clustering results.

\section{Experiments}\label{sec:experiment}
To validate the effectiveness of BURG, we conduct extensive experiments to answer the following questions: (\textbf{Q1}) Does BURG outperform current state-of-the-art methods in clustering performance on widely used datasets? (\textbf{Q2}) Do both types of consistency positively contribute to performance? (\textbf{Q3}) Does BURG effectively reduce the discrepancy between recovered data and true data? (\textbf{Q4}) How do the main hyperparameters affect BURG's performance?

\subsection{Experimental Settings}\label{sec41}
\subsubsection{Datasets and Implementation Details}
Six commonly-used multi-view datasets are adopted to validate the BURG.
\textbf{CUB}\footnote{http://www.vision.caltech.edu/visipedia/CUB-200.html} is a bird classification dataset using the first 10 categories, with two views: GoogLeNet\cite{szegedy2015going} deep visual features and doc2vec\cite{le2014distributed} textual representations.
\textbf{HandWritten}\footnote{https://archive.ics.uci.edu/ml/datasets/Multiple+Features} includes digits '0' to '9', with 200 samples per class, using six views: profile correlations, Fourier coefficients, Karhunen-Loeve coefficients, morphological, pixel features, and Zernike moments.
\textbf{CiteSeer}\footnote{https://linqs-data.soe.ucsc.edu/public/lbc/citeseer.tgz} is a graph dataset with 3,312 samples in 6 categories. We use cites, content, inbound, and outbound as four view features instead of the adjacency graph.
\textbf{Animal}\cite{lampert2013attribute} consists of 50 animal categories with 10,158 samples, using DECAF\cite{krizhevsky2012imagenet} and VGG19\cite{simonyan2015very} deep features as two views.
\textbf{Reuters}\footnote{https://archive.ics.uci.edu/ml/datasets.html} is a text dataset with 18,758 samples in 6 categories, using documents in German, English, French, Italian, and Spanish as views.
\textbf{YouTubeFace10}\cite{huang2023fast} contains 10 person categories with 38,654 samples, using features from video frames: LBP, HOG, GIST, and Gabor.

We evaluate all models in our experiments using the three widely adopted clustering metrics of ACC , NMI and ARI. We adopt Adam optimizer with the learning rate of 0.0003 for all datasets. The epochs of the three stages are set to 200, 30, and 20 respectively. The batch size of the first two stages is 128, and it is increased to 512 in the last stage. Our model is implemented based on PyTorch 2.1.0 and trained on a desktop computer configured with NVIDIA GeForce RTX 3090 and 64G RAM.



\subsubsection{Compared Methods}
We compare BURG with nine state-of-the-art DIMVC methods. Concretely, 
\textbf{GP-MVC}\cite{wang2021generative} employs CycleGAN for stable cross-view generation through cyclic training. 
\textbf{COMPLETER}\cite{lin2021completer} enhances representation consistency via contrastive learning and mutual information maximization. 
\textbf{DCP}\cite{lin2022dual} combines mutual information maximization with cross-entropy minimization for consistent representations. 
\textbf{SURE}\cite{yang2022robust} introduces a noise-robust contrastive loss to mitigate false negatives. 
\textbf{DIMVC}\cite{xu2022deep} avoids imputation by using individual view embedding with EM-like optimization. 
\textbf{APADC}\cite{xu2023adaptive} presents an imputation-free approach aligning distributions in a common space. 
\textbf{SMILE}\cite{zeng2023semantic} discovers invariant semantic distributions without paired samples. 
\textbf{DIVIDE}\cite{lu2024decoupled} reduces false negatives using high-order random walks to identify in-cluster samples. 
\textbf{ICMVC}\cite{chao2024incomplete} extracts complementary information guided by high confidence.
In these methods, only DIMVC and APADC are recovery-free approaches, while the others are recovery-based methods.

\begin{table*}[!htbp]
\centering
\resizebox{0.90\textwidth}{!}{
\begin{tabular}{|c|c|ccc|ccc|ccc|ccc|}
\hline
\multirow{2}{*}{} & Missing Rates & \multicolumn{3}{c|}{0.1} & \multicolumn{3}{c|}{0.3}                                                     & \multicolumn{3}{c|}{0.5} & \multicolumn{3}{c|}{0.7} \\ \cline{2-14}
                  & Metrics       & \multicolumn{1}{c}{ACC(\%)} & \multicolumn{1}{c}{NMI(\%)} & ARI(\%) &       \multicolumn{1}{c}{ACC(\%)} & \multicolumn{1}{c}{NMI(\%)} & ARI(\%) & \multicolumn{1}{c}{ACC(\%)} & \multicolumn{1}{c}{NMI(\%)} & ARI(\%) & \multicolumn{1}{c}{ACC(\%)} & \multicolumn{1}{c}{NMI(\%)} & ARI(\%) \\ \hline
\multirow{10}{*}{\rotatebox{90}{\textbf{CUB}}}     
        & GP-MVC             & \underline{67.33} & 65.57 & 50.15 & 55.00 & 60.26 & 42.98 & \underline{56.33} & 54.90 & \underline{39.03} & 41.33 & 46.24 & 24.97 \\
        & COMPLETER          & 52.25 & 65.06 & 46.54 & 49.33 & 61.76 & 44.42 & 41.84 & 49.61 & 32.46 & 19.22 & 14.84 & 5.46 \\
        & DCP                & 61.25 & 66.28 & 47.80 & 56.50 & 59.08 & \underline{45.38} & 32.08 & 35.71 & 16.92 & 28.25 & 25.14 & 13.65 \\
        & SURE               & 67.04 & 63.12 & 50.12 & 53.80 & 46.47 & 33.14 & 51.03 & 43.76 & 25.33 & 40.20 & 31.29 & 18.36 \\
        & DIMVC              & 60.98 & 60.77 & 44.86 & \textbf{58.26} & 50.66 & 36.82 & 42.76 & 37.20 & 30.37 & 37.58 & 27.64 & 21.63 \\
        & APADC              & 46.33 & 41.20 & 28.99 & 34.83 & 33.21 & 20.28 & 30.94 & 28.62 & 14.06 & 29.28 & 24.35 & 9.35 \\
        & SMILE              & 62.83 & \underline{67.99} & 42.80 & 55.80 & 59.08 & 37.16 & 54.33 & \underline{56.16} & 31.29 & \underline{44.17} & \underline{48.62} & 27.34 \\
        & DIVIDE             & 62.50 & 59.75 & 44.47 & 47.83 & 50.84 & 34.08 & 45.67 & 46.04 & 28.20 & 34.50 & 34.51 & 21.54 \\
        & ICMVC              & 62.83 & 66.54 & \underline{50.18} & 57.00 & \underline{60.58} & 43.13 & 45.67 & 50.79 & 33.48 & 42.19 & 45.80 & \underline{30.71} \\
        & \textbf{BURG}      & \textbf{71.50} & \textbf{71.60} & \textbf{60.48} & \underline{58.00} & \textbf{61.85} & \textbf{45.86} & \textbf{57.33} & \textbf{58.89} & \textbf{45.12} & \textbf{53.83} & \textbf{56.58} & \textbf{45.12} \\ \hline
        
\multirow{10}{*}{\rotatebox{90}{\textbf{HandWritten}}}     
        & GP-MVC             & 91.20 & 83.52 & \underline{81.56} & \underline{86.30} & 76.67 & 72.51 & 75.40 & 66.34 & 57.86 & 69.55 & 61.80 & 51.41 \\
        & COMPLETER          & 70.50 & 75.46 & 62.42 & 63.45 & 63.80 & 49.40 & 50.57 & 50.80 & 38.64 & 35.65 & 32.12 & 21.23 \\
        & DCP                & 75.32 & 77.24 & 67.39 & 75.57 & 78.36 & 69.34 & 74.20 & \underline{72.34} & 59.83 & 62.40 & 60.60 & 40.79 \\
        & SURE               & 72.40 & 73.62 & 62.19 & 70.21 & 71.15 & 60.64 & 65.59 & 63.28 & 53.23 & 62.95 & 54.59 & 44.46 \\
        & DIMVC              & 81.99 & 76.69 & 61.88 & 80.90 & 75.66 & 61.57 & 68.77 & 65.53 & 46.97 & 67.35 & 65.92 & 46.14 \\
        & APADC              & 89.88 & 84.94 & 76.45 & 74.63 & 72.83 & 61.89 & 70.35 & 68.41 & 54.77 & 68.92 & \underline{66.51} & 50.99 \\
        & SMILE              & 87.95 & 82.66 & 70.25 & 85.40 & \underline{83.45} & 68.92 & 71.05 & 65.30 & 55.07 & 70.20 & 60.69 & \underline{54.20} \\
        & DIVIDE             & 91.85 & 86.77 & 78.61 & 78.75 & 65.24 & 58.01 & 75.45 & 61.56 & 52.02 & \underline{74.75} & 57.11 & 50.42 \\
        & ICMVC              & \underline{92.55} & \underline{87.56} & 80.00 & 83.75 & 82.17 & \underline{73.70} & \underline{75.50} & 69.85 & \underline{60.04} & 68.25 & 53.69 & 43.28 \\
        & \textbf{BURG}   & \textbf{95.70} & \textbf{90.83} & \textbf{90.66} & \textbf{95.30} & \textbf{89.74} & \textbf{89.83} & \textbf{94.30} & \textbf{88.36} & \textbf{87.76} & \textbf{79.75} & \textbf{84.27} & \textbf{75.30} \\ \hline
        
\multirow{10}{*}{\rotatebox{90}{\textbf{CiteSeer}}}     
        & GP-MVC             & 23.28 & 13.96 & 1.85 & 22.16 & 13.07 & 1.54 & 20.89 & 7.81 & 0.94 & 20.56 & 7.10 & 0.55 \\
        & COMPLETER          & 28.08 & 11.16 & 5.47 & 27.05 & 11.99 & 4.91 & 24.43 & 8.93 & 2.21 & 22.31 & 5.03 & 1.64 \\
        & DCP                & 39.89 & 20.64 & 9.74 & 38.95 & 18.40 & 9.09 & 33.91 & 11.17 & 4.99 & 31.10 & 9.21 & 2.97 \\
        & SURE               & 32.61 & 15.18 & 8.76 & 30.59 & 11.22 & 6.04 & 28.77 & 8.89 & 2.65 & 24.55 & 4.72 & 0.42 \\
        & DIMVC              & 41.27 & 25.46 & 8.10 & 37.77 & 18.65 & 4.56 & 33.21 & 12.97 & 3.87 & 32.19 & 10.90 & 2.83 \\
        & APADC              & 37.44 & 23.96 & 11.97 & 36.93 & 19.07 & 8.93 & 30.10 & 17.81 & 6.03 & 27.81 & 12.10 & \underline{5.57} \\
        & SMILE              & 41.73 & 25.34 & 7.76 & 37.77 & 18.53 & 4.59 & 36.59 & 14.56 & 4.50 & 32.52 & 10.87 & 2.91 \\
        & DIVIDE             & 43.36 & \underline{25.54} & \underline{14.00} & 38.01 & 18.57 & 8.42 & 35.75 & 14.44 & 4.74 & 30.22 & 10.11 & 4.14 \\
        & ICMVC              & \underline{44.19} & 24.69 & 13.47 & \underline{40.40} & \underline{20.02} & \underline{11.43} & \underline{40.22} & \underline{20.53} & \underline{10.09} & \underline{33.85} & \underline{12.35} & 5.22 \\
        & \textbf{BURG}   & \textbf{45.38} & \textbf{27.92} & \textbf{14.61} & \textbf{45.41} & \textbf{20.32} & \textbf{11.60} & \textbf{47.43} & \textbf{21.70} & \textbf{12.36} & \textbf{35.69} & \textbf{14.61} & \textbf{7.56} \\ \hline

\multirow{10}{*}{\rotatebox{90}{\textbf{Animal}}}     
        & GP-MVC             & 44.01 & 54.76 & 30.50 & 43.38 & 53.61 & 29.19 & 39.29 & 48.12 & 24.02 & 38.58 & 47.57 & 22.51 \\
        & COMPLETER          & 29.81 & 46.66 & 17.83 & 24.54 & 39.20 & 15.30 & 20.84 & 34.82 & 13.58 & 15.78 & 30.62 & 10.54 \\
        & DCP                & 28.86 & 47.14 & 20.00 & 28.14 & 47.04 & 18.84 & 26.34 & 44.00 & 16.25 & 24.79 & 43.62 & 14.98 \\
        & SURE               & 38.23 & 49.14 & 25.48 & 36.38 & 48.92 & 24.08 & 35.94 & 47.06 & 23.70 & 34.03 & 44.34 & 21.55 \\
        & DIMVC              & 48.29 & 60.20 & 35.79 & \underline{47.74} & \underline{59.72} & \underline{37.71} & \underline{46.42} & \underline{56.31} & 33.46 & \textbf{45.27} & 53.06 & 28.31 \\
        & APADC              & 30.23 & 48.44 & 25.01 & 29.15 & 46.72 & 23.31 & 28.79 & 45.51 & 22.72 & 27.33 & 43.39 & 20.79 \\
        & SMILE              & 48.46 & 60.78 & \underline{39.48} & 46.85 & 57.65 & 36.80 & 45.31 & 54.67 & \underline{34.10} & 40.90 & \underline{53.83} & \underline{30.70} \\
        & DIVIDE             & \underline{49.18} & \underline{61.64} & 38.58 & 42.70 & 53.40 & 30.45 & 38.51 & 49.91 & 26.37 & 34.42 & 45.42 & 20.89 \\
        & ICMVC              & 45.10 & 60.73 & 33.62 & 41.17 & 57.19 & 31.55 & 38.34 & 54.30 & 28.76 & 34.67 & 50.71 & 25.14 \\
        & \textbf{BURG}   & \textbf{49.24} & \textbf{62.84} & \textbf{41.60} & \textbf{49.11} & \textbf{61.89} & \textbf{40.28} & \textbf{48.61} & \textbf{59.08} & \textbf{39.71} & \underline{44.55} & \textbf{55.62} & \textbf{34.80} \\ \hline
\multirow{10}{*}{\rotatebox{90}{\textbf{Reuters}}}     
        & GP-MVC             & 32.39 & 19.47 & 16.91 & 30.96 & 17.20 & 13.96 & 27.48 & 12.39 & 8.65 & 23.73 & 10.45 & 6.44 \\
        & COMPLETER          & 39.54 & 21.43 & 12.81 & 37.59 & 20.85 & 11.77 & 32.76 & 13.03 & 8.46 & 26.97 & 8.32 & 4.81 \\
        & DCP                & \underline{43.51} & 23.95 & 16.96 & 41.89 & 22.95 & 15.97 & 40.19 & 21.64 & 14.42 & 39.03 & 20.91 & 13.23 \\
        & SURE               & 41.32 & 25.66 & 13.32 & 40.44 & 22.40 & 10.59 & 37.63 & 21.28 & 7.58 & 37.03 & 18.21 & 6.63 \\
        & DIMVC              & 43.47 & 24.90 & 18.33 & \underline{42.51} & \underline{24.42} & 17.94 & \underline{41.91} & \underline{21.72} & \underline{17.76} & \underline{40.81} & 19.99 & 13.81 \\
        & APADC              & 39.88 & 21.51 & 20.11 & 36.80 & 20.14 & 18.76 & 31.72 & 17.42 & 15.79 & 29.99 & 16.48 & \underline{15.08} \\
        & SMILE              & 42.39 & 23.78 & 15.43 & 39.03 & 22.51 & 15.02 & 36.08 & 20.51 & 14.04 & 35.58 & 18.98 & 12.03 \\
        & DIVIDE             & 40.90 & 23.64 & 18.96 & 39.59 & 22.55 & 14.78 & 38.01 & 21.62 & 15.81 & 37.86 & 20.98 & 15.01 \\
        & ICMVC              & 42.95 & \underline{26.28} & \underline{20.72} & 40.89 & 23.89 & \underline{19.25} & 39.48 & 21.01 & 16.65 & 37.64 & \underline{22.81} & 14.17 \\
        & \textbf{BURG}   & \textbf{44.01} & \textbf{26.39} & \textbf{21.15} & \textbf{42.86} & \textbf{26.38} & \textbf{20.22} & \textbf{42.97} & \textbf{24.77} & \textbf{19.61} & \textbf{41.84} & \textbf{23.63} & \textbf{20.60} \\ \hline
        
\multirow{10}{*}{\rotatebox{90}{\textbf{YouTubeFace10}}}     
        & GP-MVC             & 76.81 & \underline{82.44} & \underline{72.73} & 73.68 & \underline{80.70} & \underline{70.81} & 65.98 & 73.28 & 53.52 & 63.69 & 71.07 & 50.11 \\
        & COMPLETER          & 62.73 & 60.79 & 39.82 & 51.89 & 57.72 & 35.67 & 52.39 & 57.42 & 29.99 & 49.05 & 53.92 & 24.52 \\
        & DCP                & 64.46 & 72.97 & 56.48 & 59.39 & 59.65 & 40.50 & 57.62 & 60.50 & 27.54 & 51.39 & 56.29 & 27.29 \\
        & SURE               & 72.65 & 72.70 & 55.95 & 65.39 & 66.29 & 40.60 & 52.90 & 58.76 & 32.12 & 49.76 & 56.12 & 26.04 \\
        & DIMVC              & 74.94 & 79.34 & 68.96 & 73.87 & 76.42 & 63.97 & \underline{68.96} & \underline{75.44} & 60.29 & 61.77 & 61.32 & 54.59 \\
        & APADC              & 56.75 & 47.16 & 40.36 & 48.09 & 39.66 & 31.89 & 39.71 & 33.97 & 23.95 & 29.23 & 25.55 & 14.43 \\
        & SMILE              & 73.50 & 78.04 & 66.55 & 72.47 & 77.32 & 64.37 & 67.91 & 73.86 & \underline{60.68} & 64.74 & 66.79 & 41.52 \\
        & DIVIDE             & \underline{77.67} & 81.11 & 69.44 & \underline{75.07} & 77.86 & 63.33 & 68.92 & 75.34 & 60.50 & \underline{67.12} & \underline{73.10} & \underline{57.22} \\
        & ICMVC              & 68.90 & 74.17 & 60.06 & 67.44 & 74.59 & 60.02 & 62.76 & 65.44 & 53.03 & 58.00 & 58.67 & 47.14 \\
        & \textbf{BURG}   & \textbf{78.92} & \textbf{83.68} & \textbf{72.97} & \textbf{80.69} & \textbf{82.47} & \textbf{72.79} & \textbf{70.41} & \textbf{76.08} & \textbf{61.17} & \textbf{67.86} & \textbf{75.00} & \textbf{59.23} \\ \hline
        
\end{tabular}}
\caption{The clustering results on six multi-view benchmarks under different missing rates. The optimal results are highlighted in \textbf{bold}, while the second-best are \underline{underlined}.}
\label{tab:performance}
\end{table*}

\subsection{Performance Comparisons(Q1)}\label{sec42}
We employ three classic metrics to quantify the clustering performance of BURG compared to the other nine methods across all datasets at different missing rates (MR = 0.1, 0.3, 0.5, 0.7) in Table \ref{tab:performance}. From the results presented in these tables, we have the following observations:
\begin{itemize}
\item Overall, whether under low or high missing rates, our proposed BURG demonstrates strong superiority in clustering performance compared to both recovery-based and recovery-free methods. Particularly at higher missing rates, such as MR=0.5, from CUB to YouTubeFace10, BURG outperforms the second-best methods by 1.00\%, 18.80\%, 7.21\%, 2.19\%, 1.06\%, and 1.45\% respectively, showing more pronounced advantages.
\item BURG exhibits remarkable performance across diverse datasets: On HandWritten with six views, it achieves over 90\% in all metrics at MR=0.1; On CiteSeer, it outperforms GCN-based ICMVC without utilizing adjacency graphs; On deep feature datasets (Animal) and text datasets (Reuters), the advantages are relatively smaller due to their inherent strong representations. Notably, on the largest dataset YouTubeFace10, BURG maintains optimal performance, achieving over 80\% in both ACC and NMI at MR=0.3 (ACC=80.69\%, NMI=82.47\%), demonstrating excellent scalability.
\end{itemize}

\subsection{Model Analysis(Q2)}\label{sec43}

\subsubsection{Ablation Study}

\begin{table}[h]
    \centering
    \resizebox{0.48\textwidth}{!}{
    \begin{tabular}{ccccc}
        \toprule
        Dataset & Dual Consistency & ACC(\%) & NMI(\%) & ARI(\%) \\
        \midrule
        \multirow{3}{*}{HandWritten} 
        & None     & 52.65 & 58.15 & 45.45 \\
        & NAC Only & 75.90 & 67.36 & 46.57 \\
        & PC Only  & 75.80 & 81.19 & 70.60 \\
        & NAC + PC & \textbf{94.30} & \textbf{88.36} & \textbf{87.76} \\
        \midrule
        \multirow{3}{*}{Animal} 
        & None     & 28.44 & 38.69 & 19.34 \\
        & NAC Only & 47.32 & 57.70 & 33.93 \\
        & PC Only  & 46.64 & 58.70 & 38.27 \\
        & NAC + PC & \textbf{48.61} & \textbf{59.08} & \textbf{39.71} \\
        \bottomrule
    \end{tabular}}
    \caption{Effect of the dual-consistency guided recovery strategy on HandWritten and Animal at missing rate of 0.5. The best results are highlighted in \textbf{bold}.}
    \label{tab:ablation}
\end{table}

We performed ablation studies on HandWritten and Animal datasets to evaluate our proposed neighbor-aware consistency (NAC) and prototypical consistency (PC) mechanisms. Table \ref{tab:ablation} compares four configurations at 0.5 missing rate: no consistency strategies (None), single strategies (NAC Only/PC Only), and the full model (NAC+PC).

NAC maintains local topology during missing view reconstruction, reducing noise from view heterogeneity. On the Animal dataset (50 clusters), NAC alone improves ACC from 28.44\% to 48.61\%. Meanwhile, PC ensures cross-view cluster compactness by aligning recovered embeddings with global prototypes. Notably, PC generally contributes more significantly than NAC - removing PC causes substantial performance drops, as seen in HandWritten where ACC decreases from 94.30\% to 75.90\%. These findings highlight PC's crucial role in bridging semantic gaps across heterogeneous views.

\subsubsection{Comparative Verification of Dual Consistency}
\begin{figure}[t]
    \centering  
    \begin{subfigure}[t]{0.49\linewidth}  
        \includegraphics[width=\linewidth]{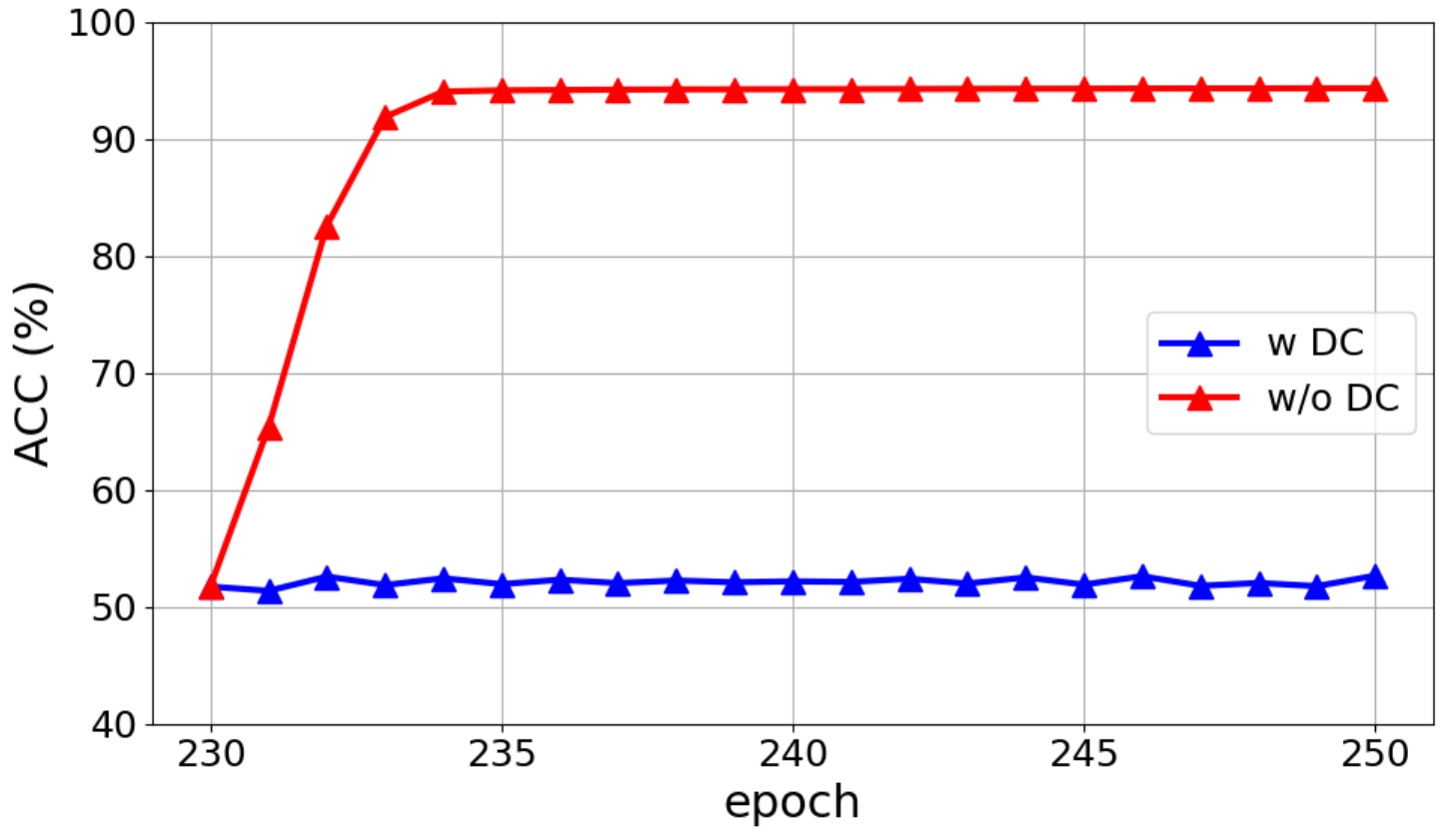}  
        \caption{HandWritten}  
    \end{subfigure}  
    \begin{subfigure}[t]{0.49\linewidth}  
    \includegraphics[width=\linewidth]{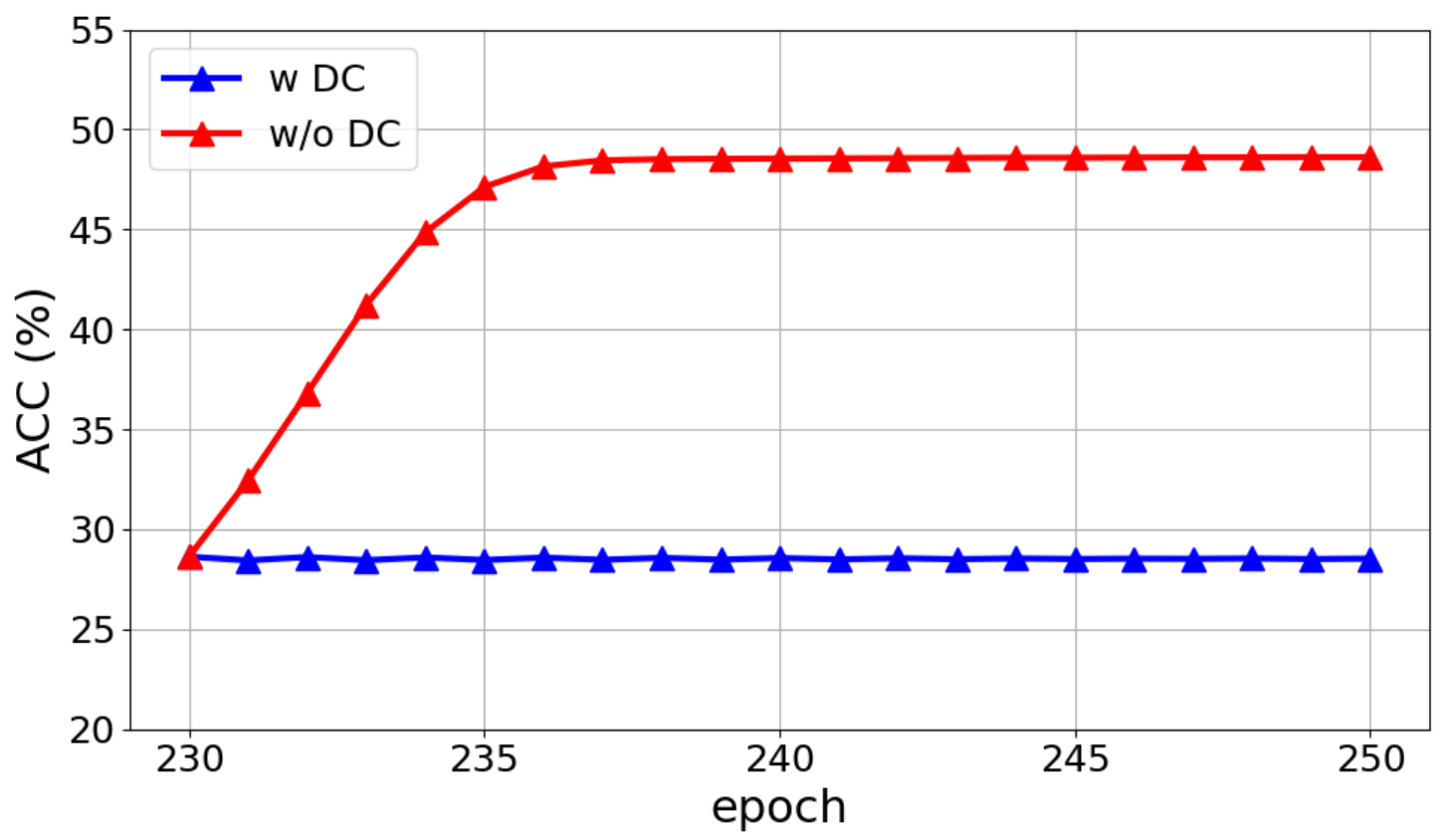}  
        \caption{Animal}  
    \end{subfigure} 
    \caption{}
\label{last20epoch}
\end{figure}

To ensure that the positive impact on clustering performance stems from our dual-consistency recovery guidance rather than simply from the additional 20 epochs of training, we present in Fig. \ref{last20epoch} the ACC curves across epochs with and without dual consistency (DC). These results clearly validate that the improvements in clustering performance are indeed attributable to our DC mechanism.

\subsubsection{Analysis of Inter-View Correlation Enhancement}
To investigate whether our proposed BURG effectively enhances the correlation consistency across different views, we evaluated its performance with varying numbers of views on HandWritten and Reuters datasets, which contain the highest number of views among all datasets. The results are presented in Table \ref{tab:diffviews}. We observe that the performance of our model demonstrates an upward trend on both datasets as the number of views increases.

\begin{table}[h]
    \centering
    \resizebox{0.48\textwidth}{!}{
    \begin{tabular}{ccccc}
        \toprule
        Dataset & View Number & ACC(\%) & NMI(\%) & ARI(\%) \\
        \midrule
        \multirow{3}{*}{HandWritten} 
        & 2V     & 72.00 & 71.23 & 60.67 \\
        & 3V & 78.15 & 77.04 & 69.14 \\
        & 4V  & 81.10 & 82.27 & 75.09 \\
        & 5V  & 92.65 & 87.56 & 85.07 \\
        & 6V & \textbf{94.30} & \textbf{88.36} & \textbf{87.76} \\  
        \midrule
        \multirow{3}{*}{Reuters} 
        & 2V     & 40.78 & 18.52 & 16.29 \\
        & 3V & 39.21 & 19.91 & 17.77 \\
        & 4V  & 42.48 & 24.07 & \textbf{22.14} \\
        & 5V & \textbf{42.97} & \textbf{24.77} & 19.61 \\
        \bottomrule
    \end{tabular}}
    \caption{Experiment with different numbers of views on HandWritten and Reuters at missing rate of 0.5. The best results are highlighted in \textbf{bold}.}
    \label{tab:diffviews}
\end{table}

\subsection{Visualization of the Discrepancy(Q3)}\label{sec45}

\begin{figure}[t]
    \centering
    \begin{subfigure}[t]{0.49\linewidth}  
        \includegraphics[width=\linewidth]{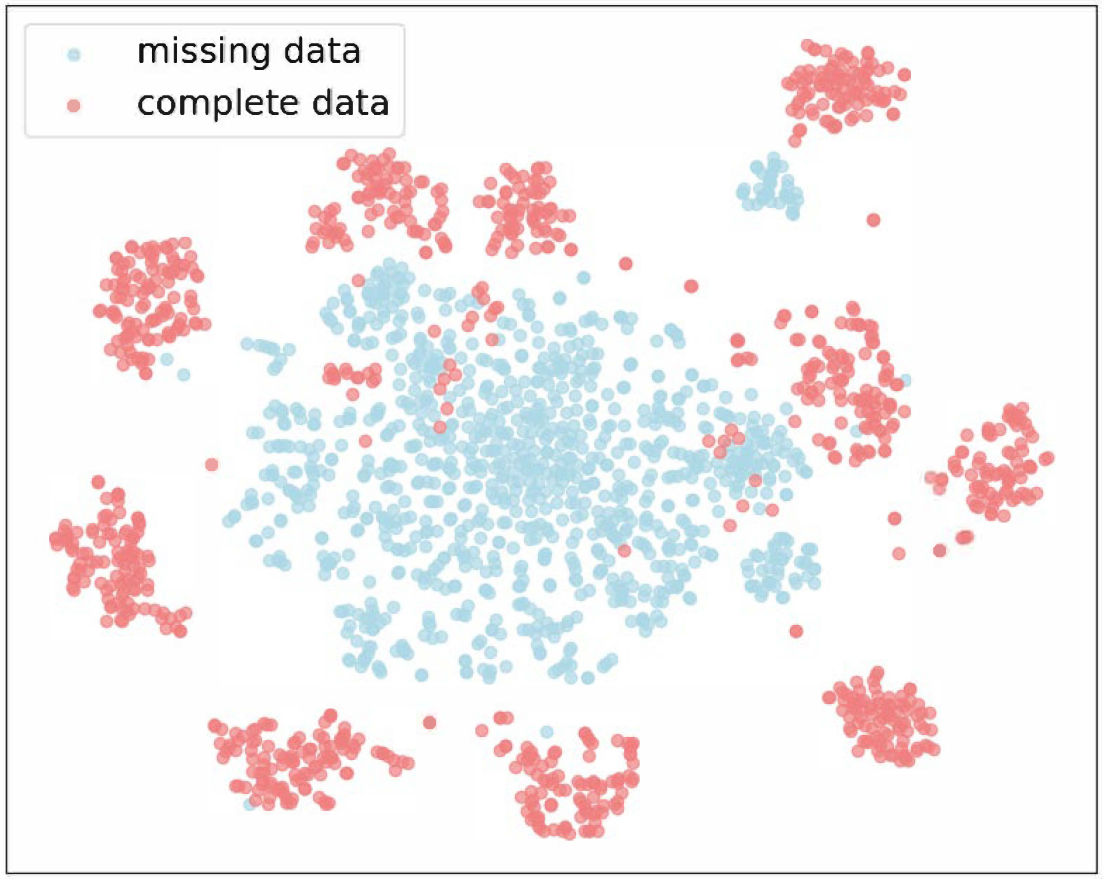}  
        \caption{Raw features}  
    \end{subfigure}  
    \begin{subfigure}[t]{0.49\linewidth}  
    \includegraphics[width=\linewidth]{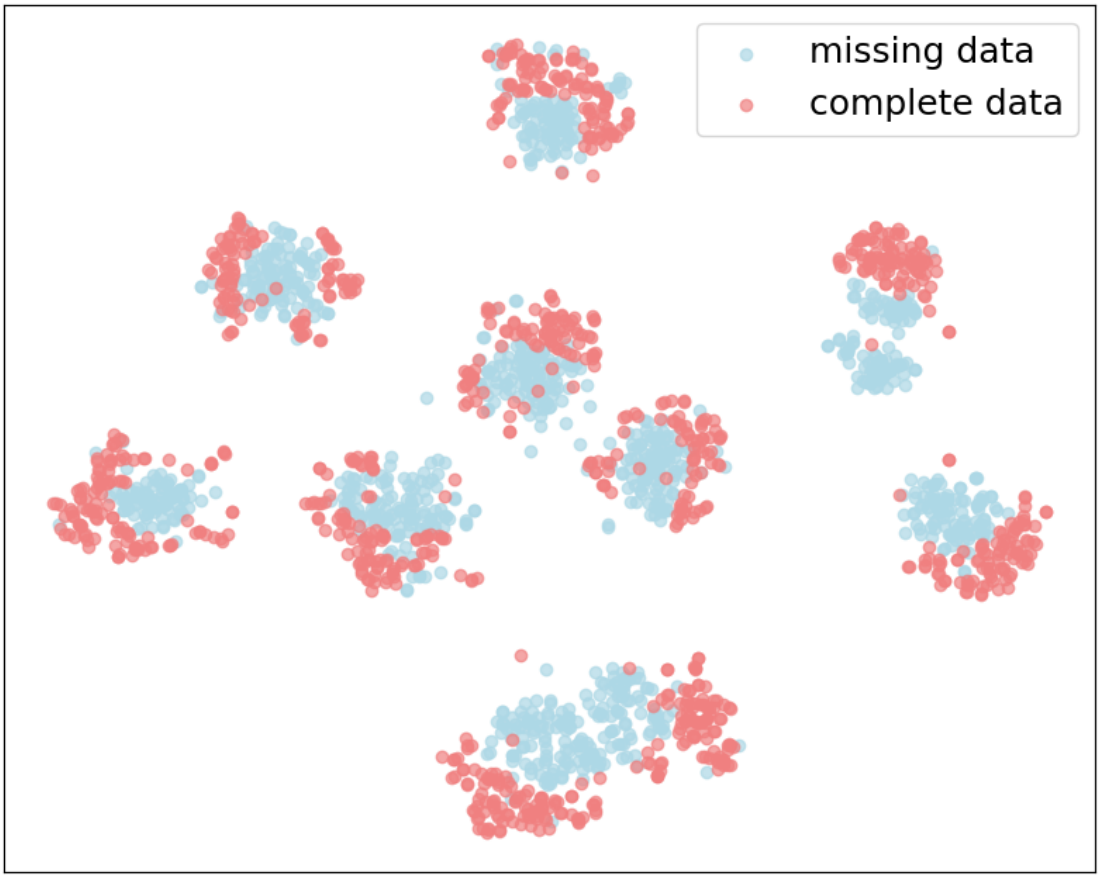}  
        \caption{After recovery}  
    \end{subfigure} 
\caption{Visualization of the discrepancy between recovered and ground-truth data on the HandWritten dataset before and after training, with a missing rate of 0.5.}
\label{tsneHW}
\end{figure}

To validate whether our proposed BURG effectively reduces the discrepancy between recovered and ground-truth data, we conducted t-SNE visualization on the HandWritten dataset under a missing rate of 0.5, as illustrated in Fig. \ref{tsneHW}. In the visualization, blue points represent samples with missing views, while red points denote samples with complete views. The visualization results strongly substantiate our motivation.

\subsection{Parameter Analysis(Q4)}\label{sec44}
\begin{figure}[t]
    \centering  
    \begin{subfigure}[t]{0.49\linewidth}  
        \includegraphics[width=\linewidth]{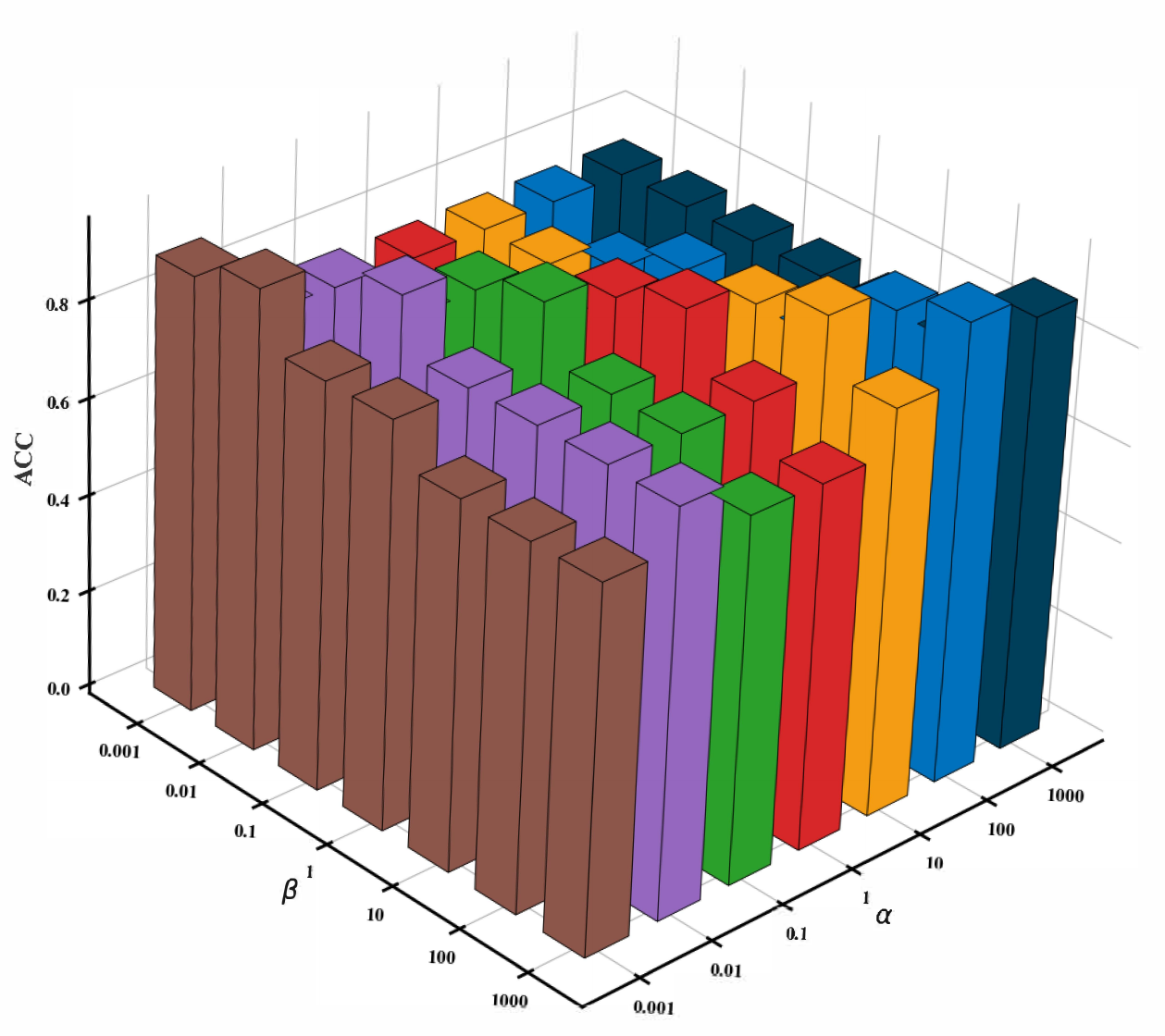}  
        \caption{HandWritten}  
    \end{subfigure}  
    \begin{subfigure}[t]{0.49\linewidth}  
    \includegraphics[width=\linewidth]{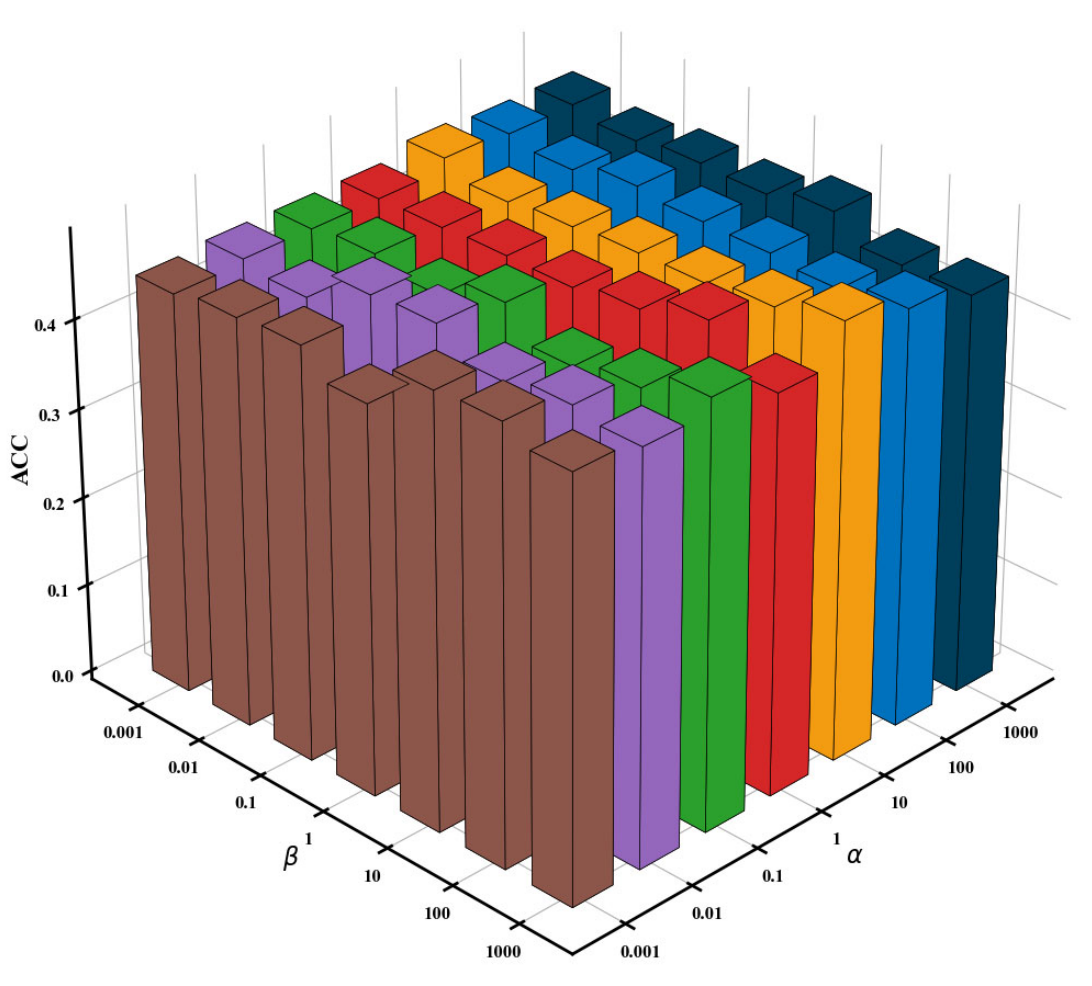}  
        \caption{Animal}  
    \end{subfigure} 
    
    \caption{Sensitivity analysis of $\alpha$ and $\beta$ for our method over HandWritten and Animal (The missing rate is 0.5).}
\label{sensitivityfig}
\end{figure}

We conducted parameter sensitivity analysis on two trade-off parameters, $\alpha$ and $\beta$, in Eq. (\ref{obj_func}) using the HandWritten and Animal datasets, as illustrated in Fig. \ref{sensitivityfig}. On the HandWritten dataset, we observed superior clustering performance when the ratio of $\alpha$ to $\beta$ was relatively large, indicating that prototypical consistency plays a dominant role. In contrast, the Animal dataset exhibited minimal fluctuations in clustering results.

\section{Conclusion}
In this paper, we propose a distribution transfer recovery method with dual consistency to restore missing instances in multi-view data. Compared to existing methods, our approach introduces two key improvements: i) utilizing flow-based generative model, it integrates the distributions of available views and transfers them to missing views, effectively predicting the distribution space of the missing views and addressing inter-view heterogeneity; ii) during the recovery process, it incorporates neighbor-aware and consistent prototypical guidance, providing rich intra-view structural information and inter-view clustering information, ensuring the distinguishability of the recovered views.

\vspace{-5pt}
\section*{Acknowledgments}
\vspace{-5pt}
This work is supported by National Science and Technology Innovation 2030 Major Project under Grant No. 2022ZD0209103. This work is supported by National Natural Science Foundation of China under Grant No. 62476281, 62441618, 62325604 and 62276271. This work is supported by National Natural Science Foundation of China Joint Found under Grant No. U24A20323.

{\small
\bibliographystyle{ieee_fullname}
\bibliography{egbib}
}

\end{document}